\documentclass{article}

\usepackage{arxiv}

\usepackage[utf8]{inputenc} 
\usepackage[T1]{fontenc}    
\usepackage{hyperref}       
\usepackage{url}            
\usepackage{booktabs}       
\usepackage{amsfonts}       
\usepackage{nicefrac}       
\usepackage{microtype}      
\usepackage{lipsum}
\usepackage{graphicx}
\graphicspath{ {./images/} }

\usepackage{algorithm}
\usepackage{algpseudocode}
\usepackage{amsmath}

\title{Entropy-Guided Sequence Weighting for Efficient Exploration in RL-Based LLM Fine-Tuning}

\author{
 Abdullah Vanlioglu \\
  SADC Department\\
  Huawei Türkiye R\&D Center\\
  Istanbul, Türkiye \\
  \texttt{abdullah.vanlioglu@huawei.com} \\
}

\begin{document}
\maketitle
\begin{abstract}
We introduce Entropy-Guided Sequence Weighting (EGSW), a novel approach that enhances the exploration-exploitation tradeoff by dynamically assigning weights to generated outputs based on their advantage and entropy for Reinforcement Learning-based Large Language Model fine-tuning. EGSW integrates entropy regularization with advantage-based weighting to balance policy updates, enabling efficient exploration in high-dimensional state spaces. By employing temperature-scaled softmax weighting over sequences, EGSW prioritizing high-reward, high-uncertainty steps while maintaining training stability. Although originally developed to improve Group Relative Policy Optimization (GRPO) during large language model (LLM) fine-tuning, EGSW is generalizable to other reinforcement learning (RL) algorithms and can be implemented in both step-wise and trajectory-wise settings. Empirical evaluations demonstrate that EGSW enhances GRPO reasoning ability, yielding improvements in sample efficiency. Future work will explore the application of EGSW to advanced RL methodologies.
\end{abstract}


\section{Introduction}
Large Language Models (LLMs) have demonstrated remarkable capabilities across various domains, from natural language understanding to code generation. However, their ability to perform complex reasoning tasks—such as multi-step mathematical problem-solving, logical inference, and structured decision-making remains an ongoing challenge. While Chain-of-Thought (CoT) \cite{wei2022chain} prompting has significantly improved reasoning by encouraging models to generate intermediate steps before producing a final answer, these improvements are limited by static prompting and do not adaptively optimize reasoning strategies based on feedback.

To address this limitation, Reinforcement Learning (RL) based fine-tuning has emerged as an effective approach for improving reasoning capabilities. By treating reasoning as a sequential decision-making process, RL enables models to iteratively refine their output strategies based on a reward signal provided by an external verifier or learned reward model. This paradigm allows LLMs to optimize their intermediate reasoning steps in a structured, feedback-driven manner. However, fine-tuning LLMs with RL presents significant challenges, particularly in high-dimensional state spaces where efficient exploration is crucial.

Exploration methods such as Monte Carlo Tree Search (MCTS), as used in AlphaZero \cite{silver2017mastering}, have been highly successful in structured domains like games. However, the computational cost of MCTS makes it impractical for LLM fine-tuning. While inference-time search strategies, such as ChatGPT-O1, improve output quality, they come with significant latency and resource constraints. On the other hand, search-free methods like DeepSeek-r1 \cite{guo2025deepseek}, which rely on Group Relative Policy Optimization (GRPO) \cite{shao2024deepseekmath}, explore the space using only policy-driven sampling, often leading to suboptimal exploration. Unlike AlphaZero, which systematically evaluates multiple trajectories via MCTS, GRPO relies on the model’s inherent sampling, making exploration inefficient, especially in high-dimensional reasoning tasks.

To overcome these challenges, we propose Entropy-Guided Sequence Weighting (EGSW), a method that improves exploration efficiency in RL-based LLM fine-tuning while maintaining computational feasibility. EGSW dynamically balances exploration and exploitation by assigning adaptive weights to generated sequences based on their entropy and advantage values. By employing a temperature-scaled softmax weighting mechanism, EGSW ensures that training prioritizes sequences with both high reward and high uncertainty, fostering a more effective learning process. Unlike MCTS-based methods, which rely on exhaustive search, EGSW provides a lightweight yet effective alternative that enhances exploration without requiring extensive search trees. Furthermore, EGSW is flexible enough to be integrated into both step-wise and trajectory-wise RL frameworks, making it broadly applicable across various RL methodologies.

This paper introduces the following key contributions:
\begin{itemize}
    \item Entropy-Guided Sequence Weighting (EGSW): A novel approach that integrates entropy with advantage-based weighting to improve exploration efficiency in RL-based fine-tuning.

    \item Improved LLM Fine-Tuning: We demonstrate that EGSW enhances reasoning performance across benchmark tasks, including MATH-500 \cite{lightman2023let}, AIME, and GPQA Diamond.

    \item Scalability and Flexibility: Unlike computationally expensive search-based methods, EGSW offers an efficient alternative that can be applied to both step-wise and trajectory-wise RL frameworks, with potential extensions to other advanced RL algorithms.
\end{itemize}

By addressing key limitations in search-based and search-free RL fine-tuning methods, EGSW provides an efficient and effective mechanism for enhancing reasoning capabilities in LLMs while maintaining practical computational constraints.

\section{Related Work}
\label{sec:related}
Balancing exploration and exploitation is a fundamental challenge in reinforcement learning (RL), especially when fine-tuning large language models (LLMs) for complex reasoning tasks. This balance is crucial to ensure that models not only utilize existing knowledge effectively but also seek out new information to improve performance.

MCTS has been effectively employed in structured domains, notably in game-playing AI such as AlphaZero, to balance exploration and exploitation through systematic state-space exploration. Recent research has adapted MCTS for LLMs to improve reasoning capabilities. For instance, the study "Generating Code World Models with Large Language Models Guided by Monte Carlo Tree Search" \cite{NEURIPS2024_6f479ea4} demonstrates that integrating MCTS with LLMs can enhance model-based RL by generating precise and efficient code representations of world-model. Similarly, "Monte Carlo Tree Search Boosts Reasoning via Iterative Preference Learning" \cite{xie2024monte} illustrates that MCTS can enhance LLM reasoning by iteratively refining model preferences. However, the computational demands of MCTS limit its practicality for large-scale LLM fine-tuning. ChatGPT-O1 addresses the exploration challenge by applying MCTS at inference time. This approach improves response quality by exploring possible response trajectories and selecting the most promising ones. While effective, inference-time MCTS introduces significant latency and computational overhead, making it unsuitable for real-time applications or large-scale deployment.

Entropy regularization has been introduced in RL to promote policy stochasticity, thereby enhancing exploration. The method involves augmenting the RL objective with an entropy term, encouraging policies that are more exploratory and robust to noise. The work "Average-Reward Reinforcement Learning with Entropy Regularization" \cite{adamczyk2025average} discusses how entropy-regularized RL leads to naturally stochastic optimal policies, improving exploration without relying on heuristic methods. Also, Soft Actor-Critic (SAC) \cite{haarnoja2018soft} algorithm incorporates entropy regularization with advantage-based updates to achieve a balance between exploration and exploitation. In high-dimensional reasoning tasks, naively maximizing entropy often introduces excessive randomness rather than guiding the model toward useful and diverse reasoning strategies. Additionally, existing exploration mechanisms do not effectively prioritize reasoning steps that are both highly informative and high-reward, leading to inefficient policy updates.

Prioritized Trajectory Replay: A Replay Memory
for Data-driven Reinforcement Learning \cite{liu2023prioritized} and PTR-PPO: Proximal Policy Optimization with Prioritized Trajectory Replay \cite{liang2021ptrppo} have introduced novel approaches to improving sample efficiency and exploration in RL. Prioritized Trajectory Replay (PTR) \cite{liu2023prioritized} a method for replaying entire trajectories based on their priority, enabling more efficient use of past experiences. Similarly, PTR-PPO \cite{liang2021ptrppo} enhances PPO by incorporating prioritized trajectory replay, improving performance in complex environments. While these methods demonstrate the benefits of trajectory-level prioritization, they do not explicitly incorporate entropy regularization, which is critical for robust exploration in high-dimensional spaces.

DeepSeek V3 \cite{deepseek2024v3} adopts a search-free approach, using GRPO to fine-tune LLMs efficiently. GRPO stabilizes training by normalizing rewards within groups of trajectories and optimizing policies using per-token advantages. However, its reliance on advantage estimates often underutilizes trajectory-level uncertainty, leading to suboptimal exploration in high-dimensional state spaces.

Recent studies have further explored the exploration-exploitation dilemma in LLM fine-tuning. For example, the work "Ignore the KL Penalty! Boosting Exploration on Critical Tokens to Improve Language Model Fine-Tuning" \cite{vassoyan2025ignore} investigates the exploration dynamics of language models on tasks requiring complex reasoning, highlighting the challenges of balancing exploration with adherence to pre-trained capabilities. Additionally, "Inference-Aware Fine-Tuning for Best-of-N Sampling in Large Language Models" \cite{chow2024inference} proposes methods to fine-tune LLMs that directly optimize performance for specific inference-time strategies, addressing the non-differentiable nature of certain selection processes.

\section{Background}\label{background}
\subsection{Group Relative Policy Optimization (GRPO)}
Group Relative Policy Optimization (GRPO) is an extension of Proximal Policy Optimization (PPO) \cite{schulman2017proximal} designed to stabilize policy updates by normalizing advantage estimates over groups of samples. The key idea is to improve the robustness of policy gradients by reducing variance within groups of trajectories.

The standard PPO objective is defined as:
\begin{equation}
L_{\text{PPO}}(\theta) = \mathbb{E}_{\tau}\left[ \min\Bigl( r_t(\theta) \, A_t,\; \operatorname{clip}\bigl(r_t(\theta), 1-\epsilon, 1+\epsilon\bigr) \, A_t \Bigr) \right]
\cite{schulman2017proximal},
\label{eq:PPO}
\end{equation}
where the likelihood ratio for given question $q$ and output $a$ is expressed below:
\begin{equation}
r_t(\theta) = \frac{\pi_{\theta}(a_{i,t} \mid q, a_{i,<t})}{\pi_{\theta_{\text{old}}}(a_{i,t} \mid q, a_{i,<t})}
\cite{schulman2017proximal},
\label{eq:ratio}
\end{equation}

and \(A_t\) is the advantage at time step \(t\).

GRPO extends this framework by sampling a group of outputs for every question. Each group \(i\ \in K\) contains \(N_k\) output samples. Within each group, the advantage values are normalized to reduce variance. Specifically, for group \(i\), the normalized advantage for each step $t$ is computed as:
\begin{equation}
    \hat{A}_t^{(k)} = \frac{R_t^{(k)} - \mu^{(k)}}{\sigma^{(k)}}
    \cite{shao2024deepseekmath},
    \label{eq:norm_adv}
\end{equation}

where \(\mu^{(k)}\) and \(\sigma^{k}\) are the mean and standard deviation of the rewards in group \(k\).

In addition, GRPO incorporates a KL divergence penalty to constrain the update and prevent the new policy from deviating too far from the old policy. The KL divergence between the old and new policy is given by:
\begin{equation}
    D_{\text{KL}}\Bigl(\pi_{\theta_{\text{old}}} \,\|\, \pi_{\theta}\Bigr) = \frac{\pi_{\text{ref}}(a_{i,t} | q, a_{i,<t})}{\pi_{\theta}(a_{i,t} | q, a_{i,<t})} 
- \log \frac{\pi_{\text{ref}}(a_{i,t} | q, a_{i,<t})}{\pi_{\theta}(a_{i,t} | q, a_{i,<t})} - 1\cite{shao2024deepseekmath}.
    \label{eq:kl}
\end{equation}

With these components, the GRPO objective is expressed as:

\begin{equation}
J_{\text{GRPO}}(\theta) = \frac{1}{K} \sum_{i=1}^{K} \frac{1}{N_k}  \sum_{t=1}^{N_k} \Biggl\{ \min \Bigl( r_t(\theta) \, \hat{A}_t^{(i)}, \; \operatorname{clip}\bigl(r_t(\theta), 1-\epsilon, 1+\epsilon\bigr) \, \hat{A}_t^{(i)} \Bigr) -\beta \, \mathbb{E}_{\tau}\left[ D_{\text{KL}}\left(\pi_{\theta_{\text{old}}} \,\|\, \pi_{\theta}\right) \right] \Biggl\} \cite{shao2024deepseekmath},
\label{eq:GRPO_loss}
\end{equation}

where \(\beta\) is a coefficient that determines the strength of the KL divergence penalty.

\section{Entropy-Guided Sequence Weighting (EGSW)}

EGSW is a novel approach designed to improve the exploration-exploitation tradeoff in reinforcement learning by dynamically assigning weights to generated sequences or individual reasoning steps based on their advantage and entropy. Unlike standard methods that rely solely on advantage estimates, EGSW integrates entropy information to capture step or trajectory level uncertainty, guiding the policy update toward both high-reward and high uncertainty outputs.

Let \(A_{i, t}\) denote the advantage for step \(t\) calculated by equation \ref{eq:norm_adv}, and let \(H_{i, t}\) denote the entropy of the step \(t\). In our formulation, we compute a raw weight for sequence \(t\) as follows:
\begin{equation}
    w_{i,t}^{\text{raw}} = \exp\left(\frac{A_{i,t}+ \alpha H_{i,t}}{P}\right),
    \label{eq:raw_weight}
\end{equation}
where \(\alpha\) is a hyperparameter that scales the contribution of the entropy term, \(P\) is the temperature parameter controlling the sparsity of the weight distribution. Since GRPO updates the parameters every step, we calculate \( w_{i, t}^{\text{raw}} \) for every reasoning step \( t \). 

The entropy at step $t$, denoted as \( H_{i,t} \), is computed by summing the entropies of all actions generated at that step. If we denote the action distribution at step $t$ as $\pi_{\theta}(a_{i,t} | q, a_{i,<t})$, the entropy for step $t$ can be formulated as:

\begin{equation}
    H_{i,t} = - \sum_{a \in \mathcal{A}} \pi_{\theta}(a | q, a_{i,<t}) \log \pi_{\theta}(a | q, a_{i,<t}),
    \label{eq:step_ent}
\end{equation}

where, $\pi_{\theta}(a | q, a_{i,<t})$ is the probability of selecting action $a$ given state $q$ under policy $\pi_{\theta}$ and $\mathcal{A}$ represents the action space.

For trajectory-level entropy, we can sum over all reasoning steps in the sequence:

\begin{equation}
    H_i = \sum_{t=1}^{T} H_{i,t} = - \sum_{t=1}^{T} \sum_{a \in \mathcal{A}} \pi_{\theta}(a | q, a_{i,<t}) \log \pi_{\theta}(a | q, a_{i,<t}).
    \label{eq:traj_ent}
\end{equation}

To ensure that the weights are properly scaled, we normalize the weights using a softmax function. For a batch containing \(N\) sequences, the normalized weight for sequence \(i\) is given by:
\begin{equation}
    w_{i,t} = \frac{w_{i,t}^{\text{raw}}}{\sum_{j=1}^{N} w_{j,t}^{\text{raw}}} 
        = \frac{\exp\left(\frac{A_{i,t} + \alpha H_{i,t}}{P}\right)}{\sum_{j=1}^{N} \exp\left(\frac{A_{j,t} + \alpha H_{j,t}}{P}\right)}.
    \label{eq:normalized_weight}
\end{equation}

These normalized weights \(w_i\) are then used to reweight the policy gradient update. By normalizing the weights, we ensures that the overall gradient scale remains similar to that of standard policy gradient updates, thereby preserving training stability. In EGSW policy gradient method, the update is given by:

\begin{equation}
\nabla_\theta \mathcal{J}_{\text{EGSW}}(\theta) = \\
\frac{1}{K} \sum_{i=1}^{K} \frac{1}{N_{k}} \sum_{t=1}^{N_{k}} w_{i,t} \Bigg[ \hat{A}_{i,t} + \beta \left( \frac{\pi_{\text{ref}}(a_{i,t}|q, a_{i,<t})}{\pi_{\theta}(a_{i,t}|q, a_{i,<t})} - 1 \right) \Bigg] \nabla_\theta \log \pi_\theta(a_{i,t}|q, a_{i,<t}).
\label{eq:egsw_update}
\end{equation}


The pseudocode of the algorithm is given below in Algorithm \ref{alg:egsw}.

\begin{algorithm}[H]
\caption{Entropy-Guided Sequence Weighting (EGSW)}\label{alg:egsw}
\begin{algorithmic}[1]
\Require policy $\pi_\theta$, reward function $R_{\phi}$, task prompts $B$, Hyperparameters: temperature $P$, entropy coefficient $\beta$,
\State initialize policy $\pi_{\theta} \leftarrow \pi_{init}$ 
\For{each training iteration}
    \State update reference policy $\pi_{ref} \leftarrow \pi_{\theta}$
    \For{each step}
        \State Sample a batch $B_d$ from $B$
        \State Update the old policy model $\pi_{\theta_{\text{old}}} \leftarrow \pi_{\theta}$ 
        \State Sample K outputs by using $\pi_{\theta_{\text{old}}}$ for each question $q$ in $B_d$
        \State Compute rewards for each output from reward function $R_{\phi}$
        \State Compute the advantage estimates $A_{i,t}$ for each token
        \State Compute entropy $H_{i,t}$ for each token
        \State Compute raw weights using equation \ref{eq:raw_weight}
        \State Normalize weights with equation \ref{eq:normalized_weight}
        \State Update policy parameters $\theta$ using gradient descent with equation \ref{eq:egsw_update}
\EndFor
\EndFor
\end{algorithmic}
\end{algorithm}

\begin{table*}[t]
\centering
\caption{Benchmark scores of algorithms fine-tuned on Qwen2.5-Math-7B base model}
\label{tab:egsw_performance_1}
\begin{tabular}{lcccc}
\toprule
\textbf{Model} & \textbf{Math-500} & \textbf{AIME24} & \textbf{AIME25} & \textbf{GPQA Diamond} \\
\midrule
Qwen2.5-Math-7B & 52.4\% & 13.3\% & 0.0\% & 29.8\% \\
GRPO & 73.0 \% & 20.0 \% & 10.0 \% &  33.0 \% \\
GRPO + EGSW  & \textbf{74.1}\% & \textbf{25.0}\% & \textbf{11.65}\% &  \textbf{34.09}\% \\
\bottomrule
\end{tabular}
\end{table*}


\section{Experiments}

For fine-tune our algorithm on top of the GRPO framework, we adopt the Simple-RL Reason \cite{zeng2025simplerl} approach, which enables fine-tuning a model using a 7-billion-parameter architecture and an 8,000-example mathematical dataset. During our experiments, Qwen2.5-Math-7B \cite{yang2024qwen25mathtechnicalreportmathematical} and Qwen2.5-Math-7B-Instruct \cite{{yang2024qwen25mathtechnicalreportmathematical}} serve as the base models. We integrate our implementation using the Hugging Face open-r1 \cite{openr1}, transformers \cite{wolf-etal-2020-transformers}, and trl \cite{vonwerra2022trl} repositories.

\begin{figure}[h]
\centering
    \includegraphics[scale=0.45]{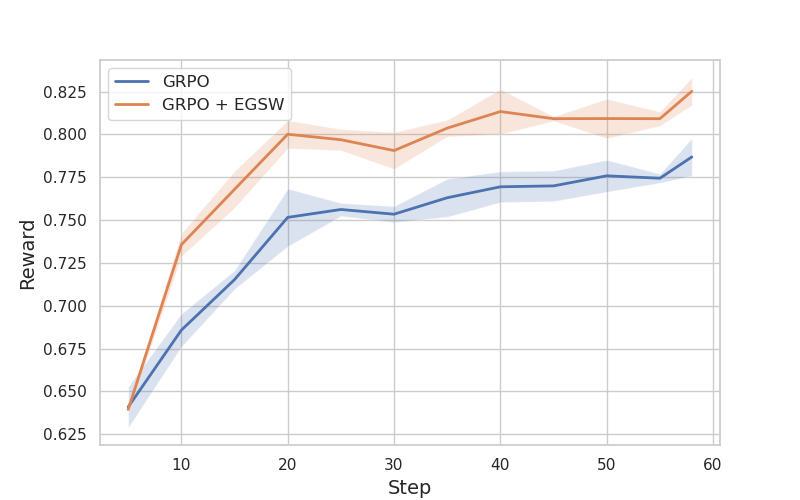}
\caption{Training reward of the methods based on Qwen2.5-Math-7B}
\label{fig:training_rew_1}
\end{figure}

Qwen2.5-Math-7B is primarily used for tasks like text completion and few-shot inference. We consider it a strong baseline for further fine-tuning to adapt specific tasks or domains. Consequently, we applied our algorithm using this model. Figure \ref{fig:training_rew_1} shows the training trend of the our algorithm (GRPO + EGSW) and GRPO. As shown in the figure our algorithm getting better score during the training. 

\begin{figure}[h]
\centering
    \includegraphics[scale=0.45]{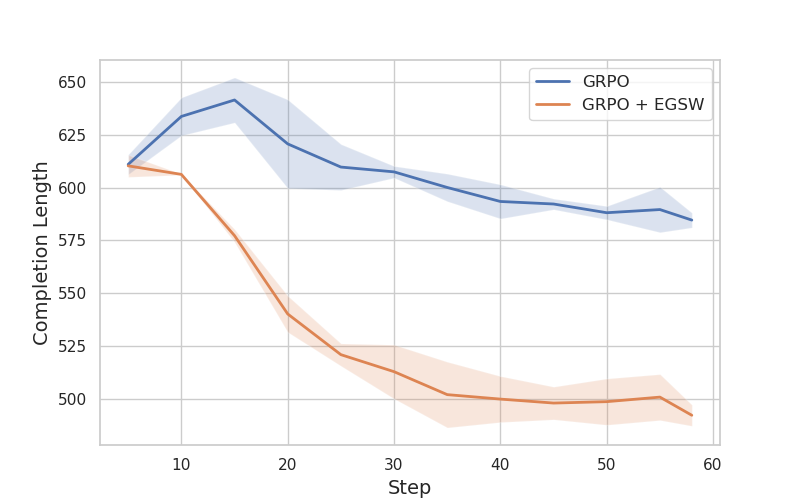}
\caption{Completion length of the methods based on Qwen2.5-Math-7B}
\label{fig:comp_len_1}
\end{figure}

We also used Qwen2.5-Math-7B-Instruct as baseline by using our algorithm. Qwen2.5-Math-7B-Instruct is model is already fine-tuned by using GRPO. 

\begin{figure}[h]
\centering
    \includegraphics[scale=0.45]{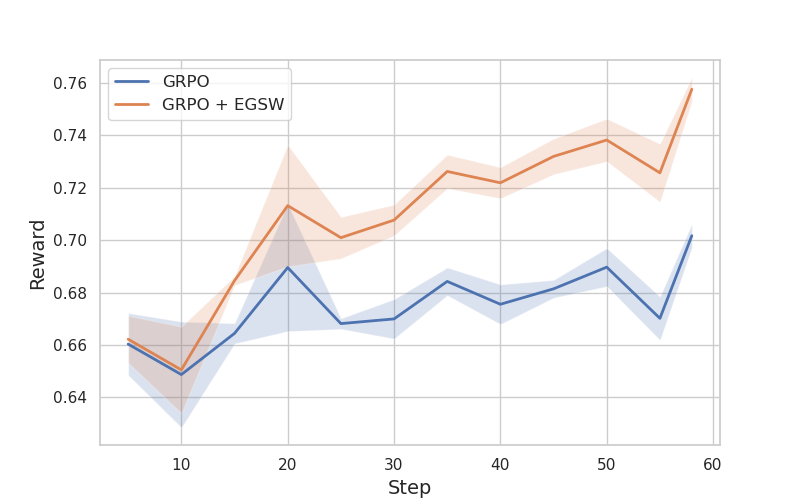}
\caption{Training reward of the methods based on Qwen2.5-Math-7B-Instruct}
\label{fig:training_rew_2}
\end{figure}

Throughout training, we observed that EGSW consistently outperformed standard Group Relative Policy Optimization in terms of reward scores. Figure \ref{fig:training_rew_2}
presents the training reward trends based on Qwen2.5-Math-7B-Instruct, comparing standard GRPO with GRPO enhanced by EGSW. The results indicate that incorporating entropy-guided weighting improves policy optimization by encouraging more effective exploration, leading to higher reward accumulation.

\begin{figure}[h]
\centering
    \includegraphics[scale=0.45]{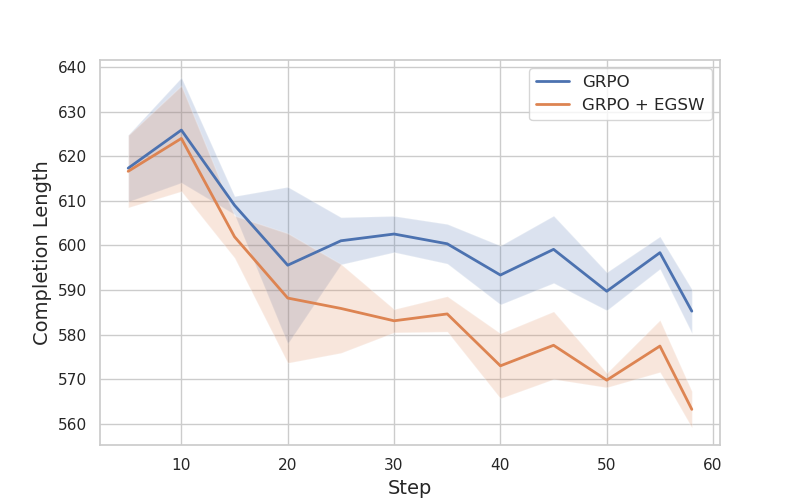}
\caption{Completion length of the methods based on Qwen2.5-Math-7B-Instruct}
\label{fig:comp_len_2}
\end{figure}

Experimental results indicate that fine-tuning Qwen2.5-Math-7B-Instruct with GRPO with 8K MATH dataset alone does not improve the model’s reasoning capability. This could be attributed to the model’s prior fine-tuning, which optimizes it for mathematical reasoning by using Qwen2.5-Math-RM. However, incorporating EGSW enhances the model’s reasoning ability. We believe this improvement stems from EGSW’s capacity to encourage better exploration. 

\begin{table*}[t]
\centering
\caption{Benchmark scores of algorithms fine-tuned on Qwen2.5-Math-7B-Instruct base model}
\label{tab:egsw_performance_2}
\begin{tabular}{lcccc}
\toprule
\textbf{Model} & \textbf{Math-500} & \textbf{AIME24} & \textbf{AIME25} & \textbf{GPQA Diamond} \\
\midrule
Qwen2.5-Math-7B-Instruct & 82.6\% & 13.33\% & 13.33\% & \textbf{38.89}\% \\
GRPO & 82.7\% & 10.0\% & 12.2\% & 32.82\% \\
GRPO + EGSW  & \textbf{84.2}\% & \textbf{14.3}\% & \textbf{17.78}\% & 32.15\% \\
\bottomrule
\end{tabular}
\end{table*}

For hyperparameter tuning, we explored several strategies. Initially, we experimented with raw entropy, adjusting the scaling coefficient between 0.15-0.50. Next, we experimented with the temperature parameter $P$ between 1 and 2. Following that, we investigated both normalized and raw entropy with an increased temperature to broaden the weight distribution.

Our results showed that the best performance for Qwen2.5-Math-7B-Instruct was achieved using normalized entropy with a scaling coefficient of $\alpha=0.8$  and setting the temperature to $P=1.8$. For Qwen2.5-Math-7B, the optimal performance was obtained with normalized entropy, using $\alpha=0.3$ and setting the temperature $P=1$. Table \ref{tab:egsw_performance_1} and Table \ref{tab:egsw_performance_2} present evaluation scores of the trained models on the Math-500, AIME24, AIME25,
and GPQA Diamond benchmarks. To obtain these scores, we trained each algorithm with three different seeds and averaged the evaluation results.

We also observed higher rewards compared to standard GRPO by using raw entropy. However, the entropy coefficient, $\alpha$, plays a crucial role in balancing exploration. During our trials, we found that setting $\alpha$ too high can lead to excessive exploration. This, in turn, causes the model to prioritize high-entropy trajectories, ultimately resulting in suboptimal performance.

Another advantage of our method is that exploration enables the model to generate fewer tokens while achieving higher rewards. Effective exploration enhances the model’s ability to focus on generating critical tokens necessary for reasoning. Figures \ref{fig:comp_len_1} and \ref{fig:comp_len_2} illustrate the completion length of the algorithms during training.

However, our algorithm is highly sensitive and requires careful hyperparameter tuning. Additionally, EGSW reduces overall gradient norm by using the weights. To ensure stable learning, it may be necessary to adjust learning-rate accordingly. 


\section{Conclusion} \label{conc}  

In this work, we introduced Entropy-Guided Sequence Weighting (EGSW), a novel approach designed to enhance the exploration-exploitation tradeoff in reinforcement learning for LLM fine-tuning. By incorporating entropy into the weighting mechanism, EGSW encourages more diverse and informative trajectories while maintaining a focus on high-reward outputs. Our method builds upon Group Relative Policy Optimization and dynamically adjusts sequence weights based on both advantage and entropy, leading to improved training stability and efficiency.  

Through extensive experiments on Qwen2.5-Math-7B and Qwen2.5-Math-7B-Instruct, we demonstrated that EGSW enhances GRPO by achieving higher reward scores and improving reasoning capabilities. Additionally, our analysis showed that EGSW facilitates efficient exploration, enabling the model to focus on generating critical tokens while reducing unnecessary completions.  

Despite its effectiveness, EGSW is highly sensitive to hyperparameter tuning, particularly the entropy scaling factor \( \alpha \) and temperature \( P \). Improper tuning may lead to excessive exploration, resulting in suboptimal performance. Future work will focus on  integrating EGSW with other reinforcement learning-based fine-tuning strategies could further enhance model performance across broader tasks.

\bibliographystyle{unsrt}  
\bibliography{references}  






\end{document}